\documentclass[]{spie}  
\usepackage[]{graphicx}
\usepackage[]{amsmath}
\usepackage[]{amssymb}
\usepackage[]{caption}
\usepackage[]{algorithm2e}
\usepackage[]{subfig} 
\usepackage[]{multirow}
\usepackage[]{array}

\DeclareMathOperator*{\argmin}{arg\,min}
\DeclareMathOperator*{\dvg}{\operatorname{div}}

\title{A Proximal Bregman Projection Approach to Continuous Max-Flow Problems Using Entropic Distances}

\author{John S.H. Baxter\supit{a,b}, Martin Rajchl\supit{a,b},
	Jing Yuan\supit{a,b}, and Terry M. Peters\supit{a,b}
\skiplinehalf
\supit{a}Robarts Research Institute, London, Ontario, Canada; \\
\supit{b}Western University, London, Ontario, Canada
}

\authorinfo{Send correspondence to J.S.H.B.: E-mail: jbaxter@robarts.ca}

\begin{document} 
  \maketitle 

\begin{abstract}
One issue limiting the adaption of large-scale multi-region segmentation is the sometimes prohibitive memory requirements. This is especially troubling considering advances in massively parallel computing and commercial graphics processing units because of their already limited memory compared to the current random access memory used in more traditional computation. To address this issue in the field of continuous max-flow segmentation, we have developed a \textit{pseudo-flow} framework using the theory of Bregman proximal projections and entropic distances which implicitly represents flow variables between labels and designated source and sink nodes. This reduces the memory requirements for max-flow segmentation by approximately 20\% for Potts models and approximately 30\% for hierarchical max-flow (HMF) and directed acyclic graph max-flow (DAGMF) models. This represents a great improvement in the state-of-the-art in max-flow segmentation, allowing for much larger problems to be addressed and accelerated using commercially available graphics processing hardware.
\end{abstract}


\section{INTRODUCTION}
\label{sec:intro}
Multi-region segmentation involves the partitioning of an image into multiple disjoint regions, each of which competes for space in the image. The rationale behind such a partitioning in an optimization-based manner is wide and multifaceted. In medical imaging, this approach can be used for segmentation and the identification of anatomy of interest from medical datasets. In computer vision, these approaches can be used for decomposing a scene into multiple regions \cite{beaulieu_hierarchy_1989, delong2009globally} or reconstructing images heavily polluted with noise \cite{bae_fast_2011}. 

Binary graph-cuts \cite{boykov_interactive_2001} were arguably the first instantiation of this approach, but have been necessarily limited, with no inherent extensibility. The binary model in a continuous space has been addressed by Yuan et al. \cite{yuan_study_2010}, specifically, the solution to the equation:
\begin{gather*}
E(u) = \int\limits_{\Omega}(D_s(x)u(x) + D_t(x)(1-u(x))+ S(x)|\nabla u(x)|)dx \\
\mbox{ s.t. } u(x) \in \{ 0, 1 \} \mbox{ .}
\end{gather*}
The technique used in the continuous max-flow model made use of augmented Lagrangian multipliers applied to the convex relaxation of the above equation. The solution to the relaxed form could then be thresholded, yielding the optimal solution to the case of discrete label values.

The simplest model reflecting this partition is the Potts model. \cite{potts_generalized_1952} This model has been largely explored by the computer vision community, yielding a proof of the NP-hardness of such a problem and the proliferation of methods for quickly approximating optimal solutions \cite{boykov_fast_2001}. Yuan et al. \cite{yuan_continuous_2010} have also addressed the corresponding continuous optimization problem:
\begin{gather*}
E(u) = \sum\limits_{\forall L} \int\limits_{\Omega}(D_L(x)u_L(x)+ S(x)|\nabla u_L(x)|)dx \\
\mbox{ s.t. } u_L(x) \geq 0 \mbox{ and } \sum\limits_{\forall L}u_L(x) = 1 \mbox{ .}
\end{gather*}
However, the Potts model is limited in that it does not allow for regions to take on variable smoothness specific to their own requirements, or to collect into super-objects expressing more complex anatomy.

The first movement towards incorporating structure and variable regularization was the discrete Ishikawa model \cite{ishikawa_exact_2003} which required the objects being segmented to be placed in a linear order. This linear order is interpreted as a sub-set relationship, that is, each object is a subset of its `parent' object in the ordering, and a superset of its `child' object. Bae et al. \cite{bae_global_2011} extended the work on the  continuous binary min-cut problem to the continuous Ishikawa model:
\begin{gather*}
E(u) = \sum\limits_{L=0}^N \int\limits_{\Omega}(D_L(x)u_L(x)+ S(x)|\nabla u_L(x)|)dx \\
\mbox{ s.t. } u_L(x) \in \{ 0,1 \} \mbox{ and } u_{L+1}(x) \leq u_L(x)
\end{gather*}
using similar variational methods but a tiered continuous graph analogous to that used by Ishikawa \cite{ishikawa_exact_2003} in the discrete case, that is, with finite capacities on intermediate flows between labels.

The Potts and Ishikawa models allowed for immediate extensibility to multiple objects or regions, but did not extend this to a flexibility in how those objects could be arranged. Such flexibility is highly desirable in segmentation problems in which complex anatomy could be modeled a complex of simpler anatomical objects under joint regularization to encourage adjacency. This has explored in the discrete domain by Delong et al.\cite{delong2011interactive,delong2012minimizing}. An early approach to this in the continuous domain was the limited hierarchical model proposed by Rajchl et al.\cite{rajchl_interactive_2014} for myocardial scar segmentation. This hierarchical approach allowed for the myocardium and scar to be given vastly different regularization parameters than the background, necessary due to their elongated structure, without necessitating a linear order over the objects. The Hierarchical Max-Flow (HMF) method in the continuous domain was formalized and generalized by Baxter et al. \cite{baxter2014ghmf}, allowing for run-time selection of hierarchies and minimizing development time in the use of hierarchical max-flow segmentation methods.

Maintaining sub-modularity for a broader selection of models has been a focus in the discrete graph cuts community due to the optimizability via graph cuts, and Delong et al.\cite{delong2009globally} have created a framework for multi-region segmentation with more general part-whole relationships being expressible. Even more recently, the continuous max-flow approach has been further generalized in Directed Acyclic Graphical Max-Flow (DAGMF) in which any set of objects can have a regularization term and these terms simultaneously applied. \cite{baxter2014dagmf} Ultimately, this family of models is the most general possible, exhausting the class of models using only the union, intersection, subset, and superset operators from set theory. (However, more models could be thought of with different constraints and considerations above simply having multiple explicit part/whole relationships.)

Focusing on continuous max-flow models, one issue with complex models is the memory requirements. In the traditional \textit{full-flow} formulations \cite{yuan_continuous_2010,bae_global_2011,baxter2014ghmf,baxter2014dagmf}, each label variable and flow variable are explicitly represented in memory, leading to a large memory footprint, especially for three-dimensional datasets. The benefit of such an approach is, by the theory of Lagrangian multipliers, the constraints placed on the label variables were implicit, and the algorithm had guaranteed convergence to a feasible solution. This did however imply that intermediate steps were not necessarily feasible, requiring some correction to the labels to be applied in post-processing, often lost in the process of discretization. 

\section{Contributions}
The primary contribution of this paper is to consolidate the theory of Bregman proximal projections\cite{bregman1967relaxation} and entropic distance metrics to continuous max-flow segmentation with multiple regions. This will first contain an algorithm and corresponding proof of correctness for the Potts model (with variable smoothness parameters). This will then be extended to the Ishikawa, HMF, and DAGMF models demonstrating the general applicability of this method to max-flow segmentation under any label ordering structure.

\section{Convex relaxed hierarchical models and Previous Work}
\subsection{Previous Work}
Traditionally, continuous max-flow segmentation models have involved the explicit representation of every flow variable in memory, leading to larger memory consumption. Recently, this has been addressed for binary models by Bae et al. \cite{bae2015maximizing} in two stages. The first implicitly represents the source flow noting that under the flow conservation constraint, it could be expressed analytically in terms of the sink flow and divergence of the spatial flow. The second method, which we will be extending in this paper, was called the \textit{pseudo-flow} representation and implicitly represented both source and sink flows, requiring only the spatial flows and labels to be stored. In this case, the six buffers required for the full-flow version is reduced to only four buffers, a significant improvement in efficiency of memory use. 

\subsection{Pseudo-Flow Approach to the Potts Model}
We will proceed in developing a pseudo-flow model following the approach taken by Bae et al. \cite{bae2015maximizing} starting with the continuous Potts model.

Yuan et al.\cite{yuan_continuous_2010} have shown the equivalence between the convex relaxed continuous Potts model:
\begin{equation}
\begin{aligned}
\min \limits_{u(x)} & \sum\limits_{\forall L} \int\limits_{\Omega} \left( D_L(x)u_L(x)+ S(x)|\nabla u_L(x)| \right) dx \\
\mbox{ s.t. } & \forall L \, u_L(x) \geq 0 \mbox{ and } \forall x, \, \sum\limits_{\forall L}u_L(x) = 1
\end{aligned}
\label{eq:potts_original}
\end{equation}
to the primal max-flow model:
\begin{equation}
\begin{aligned}
\max \limits_{p_S(x), q(x), p(x)} & \int\limits_{\Omega} p_S(x) dx \\
\mbox{ s.t. }& \forall (L, x), \, p_L(x) \leq D_L(x) \\
& \forall L, \, |q_L(x)| \leq S_L(x) \\
& \forall L, \, \dvg q_L(x) - p_S(x) + p_L(x) = 0
\end{aligned}
\label{eq:potts_primal}
\end{equation}
through the primal dual model:
\begin{equation}
\begin{aligned}
\min \limits_{u(x)} \max \limits_{p_S(x), q(x), p(x)} & \int\limits_{\Omega} \left( p_S(x) + \sum \limits_{\forall L} u_L(x) \left( \dvg q_L(x) -p_S(x) + p_L(x) \right) \right) dx \\
\mbox{ s.t. } & \forall L, \, p_L(x) \leq D_L(x) \\
& \forall L, \, |q_L(x)| \leq S_L(x) \\
\end{aligned}
\label{eq:potts_primaldual}
\end{equation}
noting that in both the primal and the primal-dual formulations, the labeling constraints $u_L(x) \geq 0$ and $\sum_{\forall L}u_L(x) = 1$ become implicit.

Firstly, we need to show that the formulation can be addressed by developing the appropriate non-smooth pseudo-flow model:
\begin{equation}
\max \limits_{ |q_L(x)| \leq S_L(x) } \int\limits_{\Omega} \min_L \left( D_L(x) + \dvg q_L(x) \right) dx
\label{eq:potts_pseudo}
\end{equation}
which can be derived from equation \eqref{eq:potts_primaldual} as shown in the following steps:
\begin{equation*}
\begin{aligned}
&\min \limits_{u(x)} \max \limits_{p_S(x), q(x), p(x)}  \int\limits_{\Omega} \left( p_S(x) + \sum \limits_{\forall L} u_L(x) \left( \dvg q_L(x) -p_S(x) + p_L(x) \right) \right) dx \\
& \mbox{ s.t. } \forall (L, x), \, p_L(x) \leq D_L(x), \; |q_L(x)| \leq S_L(x) \\
= & \min \limits_{u(x)} \max \limits_{p_S(x), q(x), p(x)} \int\limits_{\Omega} \left( p_S(x) - \sum \limits_{\forall L} u_L(x) p_S(x) + \sum\limits_{\forall L} u_L(x) \left( \dvg q_L(x) + p_L(x) \right) \right) dx \\
& \mbox{ s.t. } \forall (L, x), \, p_L(x) \leq D_L(x), \; |q_L(x)| \leq S_L(x) \\
= & \min \limits_{u(x)} \max \limits_{q(x), p(x)} \int\limits_{\Omega} \sum\limits_{\forall L} u_L(x) \left( \dvg q_L(x) + p_L(x) \right) dx \\
& \mbox{ s.t. } \sum\limits_{\forall L} u_L(x) = 1, \; \forall L, \, p_L(x) \leq D_L(x), \; |q_L(x)| \leq S_L(x)\\
& \mbox{ (as the equivalence of equations \eqref{eq:potts_original} and \eqref{eq:potts_primaldual} guarantees the constraint $\sum_{\forall L} u_L(x) = 1$ )} \\
=& \min \limits_{u(x)} \max \limits_{q(x)} \int\limits_{\Omega} \sum\limits_{\forall L} u_L(x) \left( \dvg q_L(x) + D_L(x) \right) dx \\
& \mbox{ s.t. } \sum\limits_{\forall L} u_L(x) = 1, \; \forall L, \, u_L(x) \geq 0, \; |q_L(x)| \leq S_L(x)\\
& \mbox{ (as the equivalence of equations \eqref{eq:potts_original} and \eqref{eq:potts_primaldual} guarantees the constraint $\forall L, \, u_L(x) = 1$  } \\
& \mbox{ and $\max_{p_L(x) \leq D_L(x)}u_L(x)p_L(x) = D_L(x)$ if $u_L(x)$ is non-negative) }\\
= &\max \limits_{q(x)} \min \limits_{u(x)} \int\limits_{\Omega} \sum\limits_{\forall L} u_L(x) \left( \dvg q_L(x) + D_L(x) \right) dx \\
& \mbox{ s.t. } \sum\limits_{\forall L} u_L(x) = 1, \; \forall L, \, u_L(x) \geq 0, \; |q_L(x)| \leq S_L(x)\\
= & \max \limits_{ |q_L(x)| \leq S_L(x) } \int\limits_{\Omega} \min_L \left( D_L(x) + \dvg q_L(x) \right) dx \\
& \mbox{(since $\min_{u(x)}\sum_{\forall L}u_L(x)\left( D_L(x) + \dvg q_L(x) \right) = \min_L ( D_L(x) + \dvg q_L(x) )$ when}\\
& \mbox{ $\sum_{\forall L}u_L(x) = 1$ and $\forall L, \, u_L(x) \geq 0$)}
\end{aligned}
\end{equation*}

Now that a pseudo-flow representation is developed, we can take advantage of Bregman proximal projections\cite{bregman1967relaxation} to optimize this formula. Consider the distance between vector-valued labeling functions $u(x)$ and $v(x)$ as:
\begin{equation}
d_g(u,v) = \int\limits_\Omega \left( \sum\limits_{\forall L}u_L(x) \ln ( u_L(x) / v_L(x) ) - u_L(x) + v_L(x) \right) dx
\end{equation}
which can be verified to be a Bregman distance (when $\forall L, u_L(x) \in [0,1]$) using the entropy function:
\begin{equation*}
g(u) = \int\limits_\Omega \sum\limits_{\forall L}\left( u_L(x) \ln u_L(x) - u_L(x) \right) dx \mbox{ .}
\end{equation*}

If we consider a feasible labeling, $v(x)$, we can find another proximal labeling, $u(x)$, which has a lower energy by addressing the optimization:
\begin{equation}
u(x) = \argmin\limits_{u_L(x) \geq 0, \, \sum_{\forall L}u_L(x) = 1} \left(  \max \limits_{ |q_L(x)| \leq S_L(x) } \int\limits_{\Omega} u_L(x) \left( D_L(x) + \dvg q_L(x) \right) dx + c d_g(u,v) \right)
\label{eq:potts_bregmanopt}
\end{equation}
where $c$ is a positive constant. Using a Lagrangian multiplier on the constraint $\sum_{\forall L}u_L(x) = 1$, we can solve for $u(x)$ analytically as:
\begin{equation}
u_L(x) = \frac{v_L(x) \exp \left( - \frac{D_L(x) + \dvg q_L(x) }{c} \right)}{\sum_{\forall L'} v_{L'}(x) \exp \left( - \frac{D_{L'}(x) + \dvg q_{L'}(x) }{c} \right)}
\label{eq:potts_labelupdate}
\end{equation}
noting that this answer fulfills the constraint $u_L(x) \geq 0$ provided the same holds for $v(x)$. By letting the distance weighting parameter $c$ approach 0, $u_L(x) \to 0$ if $D_L(x) + \dvg q_L(x) \neq \min_{L'} ( D_{L'}(x) + \dvg q_{L'}(x) )$. Using that fact,
\begin{equation*}
\begin{aligned}
 & \min\limits_{u_L(x) \geq 0, \, \sum_{\forall L}u_L(x) = 1} \left(  \max \limits_{ |q_L(x)| \leq S_L(x) } \int\limits_{\Omega} u_L(x) \left( D_L(x) + \dvg q_L(x) \right) dx + c d_g(u,v) \right) \\
\to & \max \limits_{ |q_L(x)| \leq S_L(x) } \int\limits_\Omega \min\limits_{L}\left( D_L(x) + \dvg q_L(x) \right) \mbox{ as } c \to 0 &
 \end{aligned}
\end{equation*}
illustrating that the Bregman proximal method acts just as well as a smoothed version of the non-smooth pseudo-flow representation, equation \eqref{eq:potts_pseudo}.

By taking the gradient of equation \eqref{eq:potts_bregmanopt} with respect to $\dvg q_L(x)$ (with $u_L(x)$ substituted by equation \eqref{eq:potts_labelupdate}) one can derive the appropriate Chambolle iteration scheme\cite{chambolle_algorithm_2004} for maximizing \eqref{eq:potts_bregmanopt} with respect to the spatial flow variables $q(x)$:
\begin{equation}
q_L(x) \gets \operatorname{Proj}_{|q_L(x)| \leq S_L(x)} \left( q_L - c\tau \nabla \left(v_L exp \left(- \frac{D_L(x) + \dvg q_L(x)}{c} \right) \right) \right)
\label{eq:potts_flowupdate}
\end{equation}
where $\tau$ is a positive gradient descent parameter.

This yields the algorithm for optimizing the continuous Potts model using repeated Bregman proximal projections and Chambolle iterations:
\begin{algorithm}[h!]
$\forall L, \, u_L(x) \gets 1 / \#L $\;
\While{not converged}{
  $\forall L, \, u_L(x) \gets u_L(x) \exp \left( - \frac{D_L(x) + \dvg q_L(x) }{c} \right) $\;
  $\forall L, \, q_L(x) \gets \operatorname{Proj}_{|q_L(x)| \leq S_L(x)} \left( q_L(x) - c\tau \nabla u_L(x) \right)$ \;
  $a(x) \gets \sum_{\forall L}u_L(x)$\;
  $\forall L, \, u_L(x) \gets u_L(x) / a(x)$\;
}
\caption{Pseudo-Flow Algorithm: Continuous Potts Model}
\end{algorithm}

\subsection{Pseudo-Flow Approach to Directed Acyclic Graphical Max-Flow}
For the sake of generalization, we shall begin developing a pseudo-flow model for DAGMF first, applying it to the special cases of HMF and Ishikawa models. As with the Potts model, note that Baxter et al. \cite{baxter2014dagmf} proved the equivalence of the DAGMF model:
\begin{equation}
\begin{aligned}
\min\limits_{u(x)} & \sum\limits_{\forall L} \int_\Omega \left( D_L(x)u_L(x)+S_{L}(x)|\nabla u_L(x)| \right) dx \\
\mbox{ s.t. } & \forall L (u_L(x) \geq 0) \\
 &  \forall L \left(u_L(x) = \sum\limits_{ L' \in L.C }w_{(L,L')} u_{L'}(x)\right) \\
 &   u_S(x)=1
\end{aligned}
\label{eq:dagmf_original}
\end{equation}
to the primal model:
\begin{equation}
\begin{aligned}
\max\limits_{p(x), q(x)} & \int\limits_\Omega p_S(x) dx \\
\mbox{ s.t. } & p_L(x)  \leq D_L(x), \text{ where } L.C = \emptyset \\
& |q_L(x)|  \leq S_L(x) \text{ } L \neq S \\
& \dvg q_L(x) + p_L(x) - \sum_{L' \in L.P}w_{(L',L)}p_{L'}(x) = 0
\end{aligned}
\label{eq:dagmf_primal}
\end{equation}
through the primal-dual model:
\begin{equation}
\begin{aligned}
\min_{u(x)} & \max\limits_{p(x), q(x)} \int \limits_\Omega p_S(x) + \sum\limits_{\forall L} u_L(x) \left( \dvg q_L(x) + p_L(x) - \sum_{L' \in L.P}w_{(L',L)}p_{L'}(x) \right) dx \\
\mbox{ s.t. } & p_L(x)  \leq D_L(x), \text{ where } L.C = \emptyset \\
& |q_L(x)|  \leq S_L(x) \text{ } L \neq S \mbox{ .}
\end{aligned}
\label{eq:dagmf_primaldual}
\end{equation}

As with before, we will develop a non-smooth pseudo-flow representation from the primal-dual representation \eqref{eq:dagmf_primaldual}. To do so, we will develop some notation, specifically:
\begin{equation*}
\mathbb{L} = \{ L | L.C = \emptyset \} ,
\end{equation*}
\begin{equation*}
d_L(x) = \left\{ 
\begin{array}{ll}
0, & \text{ if } L = S \\
D_L(x) + \dvg q_L(x) + \sum_{L' \in L.P} w_{(L',L)}d_{L'}(x), & \text{ if } L \in \mathbb{L} \\
\dvg q_L(x) + \sum_{L' \in L.P} w_{(L',L)}d_{L'}(x), & \text{ else } \\
\end{array} \right.
\end{equation*}
and
\begin{equation*}
\begin{aligned}
W_{(A,B)} = & \sum_{p \in \operatorname{path}(A,B)} \prod_{(M,N) \in p}w_{(M,N)} \\
= & \left\{ 
\begin{array}{ll}
1, & \text{ if } A = B \\
0, & \text{ if } B \text{ is not a descendant of } A, \text{ i.e. } B \notin A.C^* \\
\sum_{\forall L \in A.C}w_{(A,L)}W_{(L,B)}, & \text{ else } \\
\end{array} \right. \mbox{ . }
\end{aligned}
\end{equation*}
$d_L(x)$ where $L \in \mathbb{L}$ represents the \textit{flow excess} of an end-label, $L$, taking into account the spatial flows for super-objects containing $L$.  $W_{(A,B)}$ where $B \in \mathbb{L}$ represents the amount of weight assigned to a super-object based on a single component, the end-label $B$. These definitions mirror the top-down process of flow excess accumulation and the bottom-up process of label accumulation in the original DAGMF algorithm\cite{baxter2014dagmf} and will play such a role in the pseudo-flow formulation. $\mathbb{L}$ gives a convenient notation for end-labels.

As with the previous method, we will develop a pseudo-flow representation which is amenable to solving through a mixture of Bregman proximal projections and Chambolle iterations. The entropic distance metric used in this formulation is:
\begin{equation}
d_g(u,v) = \int\limits_\Omega \left( \sum\limits_{\forall L \in \mathbb{L}}u_L(x) \ln ( u_L(x) / v_L(x) ) - u_L(x) + v_L(x) \right) dx
\end{equation}
which can be verified to be a Bregman distance (when $\forall L, u_L(x) \in [0,1]$) using the entropy function:
\begin{equation*}
g(u) = \int\limits_\Omega \sum\limits_{\forall L \in \mathbb{L}}\left( u_L(x) \ln u_L(x) - u_L(x) \right) dx \mbox{ .}
\end{equation*}
This may at first not look like a true distance, since if there were no constraints on the values of $u_L(x)$ where $L \notin \mathbb{L}$ it would only be a pseudo-metric. However, it must be noted that since these are constrained, any two labelings in which the values are equal for end-labels must also have equal values for non-end-labels. Since $d_g(u,v)=0$ only if $u(x)$ and $v(x)$ have the same values for end-labels, $d_g(u,v)=0$ also implies they have the sames for non-end-labels and thus is a valid distance.

Consider the non-smooth pseudo-flow formulation:
\begin{equation}
\max \limits_{ |q_L(x)| \leq S_L(x) } \int\limits_{\Omega} \min_{L \in \mathbb{L}} \left( d_L(x) \right) dx
\label{eq:dagmf_pseudo}
\end{equation}
which can be derived from equation \eqref{eq:dagmf_primaldual} as follows:
\begin{equation*}
\begin{aligned}
& \min_{u(x)} \max\limits_{p(x), q(x)} \int \limits_\Omega p_S(x) \sum\limits_{\forall L} u_L(x) \left( \dvg q_L(x) + p_L(x) - \sum_{L' \in L.P}w_{(L',L)}p_{L'}(x) \right) dx \\
& \mbox{ s.t. } \forall L \in \mathbb{L}, \, p_L(x) \leq D_L(x) \\
& \forall L, \, |q_L(x)|  \leq S_L(x) \text{ } L \neq S \\
= &\min_{u(x)} \max\limits_{p(x), q(x)} \int \limits_\Omega \left( \sum\limits_{\forall L} u_L(x) \dvg q_L(x) + \sum\limits_{\forall L \in \mathbb{L}} u_L(x) p_L(x) \right) dx \\
 &\mbox{ s.t. } \forall L \in \mathbb{L}, \, p_L(x) \leq D_L(x) \\
& \forall L, \, |q_L(x)|  \leq S_L(x) \text{ } L \neq S \\
& u_S(x) = 1, \, \forall L \notin \mathbb{L}, \, u_L(x) = \sum_{L' \in L.C} w_{(L,L')}u_{L'}(x) \\
& \mbox{ (which is guaranteed by the equivalence of the DAGMF formulation and the primal-dual} \\
& \mbox{ representation) } \\
= & \min_{u(x)} \max\limits_{q(x)} \int \limits_\Omega \left( \sum\limits_{\forall L} u_L(x) \dvg q_L(x) + \sum\limits_{\forall L \in \mathbb{L}} u_L(x) D_L(x) \right) dx \\
& \mbox{ s.t. }  \forall L, \, |q_L(x)|  \leq S_L(x) \text{ } L \neq S \\
& \sum_{\forall L \in \mathbb{L}} u_L(x) = 1, \, \forall L, \, u_L(x) \geq 0 \\
& \mbox{ (which is guaranteed by the equivalence of the DAGMF formulation and the primal-dual} \\
& \mbox{ representation) } \\
=& \min_{u(x)} \max\limits_{q(x)} \int \limits_\Omega \sum\limits_{\forall L \in \mathbb{L}} u_L(x) d_L(x) dx \\
 &\mbox{ s.t. } \forall L, \, |q_L(x)|  \leq S_L(x) \text{ } L \neq S \\
& \sum_{\forall L \in \mathbb{L}} u_L(x) = 1, \, \forall L, \, u_L(x) \geq 0 \\
= &\max\limits_{q(x)}  \min_{u(x)} \int \limits_\Omega \sum\limits_{\forall L \in \mathbb{L}} u_L(x) d_L(x) dx \\
 &\mbox{ s.t. }  \forall L, \, |q_L(x)|  \leq S_L(x) \text{ } L \neq S \\
& \sum_{\forall L \in \mathbb{L}} u_L(x) = 1, \, \forall L, \, u_L(x) \geq 0 \\
= &\max\limits_{|q_L(x)| \leq S_L(x)} \int \limits_\Omega \min_{L \in \mathbb{L}} \left(d_L(x) \right) dx \mbox{ .}\\
\end{aligned}
\end{equation*}

As with earlier, we can combine this representation with the entropic distance to yield an improved labeling $u(x)$ proximal to $v(x)$ via:
\begin{equation}
\begin{aligned}
u(x) =& \argmin_{u(x)} \max\limits_{q(x)} \int \limits_\Omega \sum\limits_{\forall L \in \mathbb{L}} u_L(x) d_L(x) dx + c d_g(u,v) \\
 &\mbox{ s.t. } \forall L, \, |q_L(x)|  \leq S_L(x) \text{ } L \neq S \\
& \sum_{\forall L \in \mathbb{L}} u_L(x) = 1, \;  \forall L, \, u_L(x) \geq 0 \text{ .}
\end{aligned}
\label{eq:dagmf_bregop}
\end{equation}
This is equivalent to that used for the Potts model \eqref{eq:potts_bregmanopt}. Therefore we can solve for $u(x)$ for all end-labels $L \in \mathbb{L}$:
\begin{equation}
u_L(x) = \frac{v_L(x) \exp \left( - \frac{d_L(x) }{c} \right)}{\sum_{\forall L' \in \mathbb{L}} v_{L'}(x) \exp \left( - \frac{d_{L'}(x)}{c} \right)}
\label{eq:potts_labelupdate}
\end{equation}
and propagate these labels upwards to get the values of $u_L(x)$ where $L \notin \mathbb{L}$ using the labeling constraints. A corollary to this is that this label update equation approaches the pseudo-flow representation as $c \to 0$.

Lastly, we must consider updating the spatial flows, once again by finding the gradient of equation \eqref{eq:dagmf_bregop} with respect to $\dvg q_L(x)$. Doing so yields the update equation with positive gradient descent parameter $\tau$:
\begin{equation}
q_L(x) \gets \left\{ \begin{array}{ll}
\operatorname{Proj}_{|q_L(x)|\leq S_L(x)}\left( q_L(x) - c\tau \nabla \left( v_L(x) \exp \left( - \frac{d_L(x)}{c} \right) \right) \right) , & \text{ if } L \in \mathbb{L} \\
\operatorname{Proj}_{|q_L(x)|\leq S_L(x)}\left( q_L(x) - c\tau \nabla \left( \sum_{L' \in \mathbb{L}} W_{(L,L')} v_{L'}(x) \exp \left( - \frac{d_L'(x)}{c} \right)\right) \right) & \text{ else }
\end{array} \right. \text{ .}
\end{equation}

All together, this framework yields the following algorithm:
\begin{algorithm}[h!]
$\forall L \in \mathbb{L}, \, u_L(x) \gets 1 /  |\mathbb{L}| $\;
\While{not converged}{
  $\forall L, \, d_L(x) \gets \dvg q_L(x)$\;
  $\forall L \in \mathbb{L}, \, d_L(x) \gets d_L(x) + D_L(x)$\;
  \For{$L$ in order $\mathbb{O}/ \{ S \} $}{
  	\For{$L' \in L.C$}{
  		$d_{L'}(x) \gets d_{L'}(x) + w_{(L,L')} d_L(x)$\;
  	}
  }
  
  $\forall L \in \mathbb{L}, \, u_L(x) \gets u_L(x) \exp \left( - \frac{d_L(x) }{c} \right) $\;
  $\forall L \in \mathbb{L}, \, d_L(x) \gets u_L(x)$\;
  $a(x) \gets \sum_{\forall L\in \mathbb{L}}u_L(x)$\;
  $\forall L\in \mathbb{L}, \, u_L(x) \gets u_L(x) / a(x)$\;
  
  $\forall L \notin \mathbb{L}, \, d_L(x) \gets 0$\;
  \For{$L$ in order $\mathbb{O}^{-1} / \{ S \}$}{
  	
  	$q_L(x) \gets \operatorname{Proj}_{|q_L(x)| \leq S_L(x)} \left( q_L(x) - c\tau \nabla d_L(x) \right)$ \;
  	\For{$L' \in L.P /  \{ S \} $}{
		$d_{L'}(x) \gets d_{L'} + w_{(L',L)} d_L(x)$\;
	}
  }
}
\caption{Pseudo-Flow Algorithm: Directed Acyclic Graph Max-Flow}
\end{algorithm}
\newline
where $\mathbb{O}$ is the top-down topological ordering starting with $S$ and ending with the labels in $\mathbb{L}$ and $\mathbb{O}^{-1}$ is the inverse thereof. Note that the labels for intermediate nodes have also become implicit, further minimizing memory load. Note that this algorithm uses the same buffer for holding the flow excess $d_L$ propogating top-down and the spatial flow fields for updating the spatial flow buffers propagating up, further reducing the number of buffers required.

This algorithm requires 5 buffers per end-label, 4 buffers per non-end-label not including the source $S$, and 1 additional buffer for accumulation purposes. This is a large improvement over the full-flow approach by Baxter et al.\cite{baxter2014dagmf} which requires 6 buffers per end-label, 7 buffers per non-end-label not including the source $S$, and 2 additional buffer related to the source node.

\subsection{Application of Pseudo-Flow to Ishikawa and other Hierarchical Models}
It should be noted that the HMF and Ishikawa models are special cases of the DAGMF model, meaning that the same algorithm could be immediately applied to them. In the case of HMF, the DAGMF algorithm is equivalent to the algorithm yielded by repeating the process of developing a pseudo-flow representation and solving it using the aforementioned steps. In this case, it may be somewhat simplified as the orderings $\mathbb{O}$ and $\mathbb{O}^{-1}$ correspond to simple bottom-up and top-down tree traversals, which can be readily implemented using recursion.

The Ishikawa model is more interesting in that it's formulation renders many of the flow variables redundant, and a separate instantiation of the pseudo-flow algorithm would be warranted. Modifications to the DAGMF pseudo-flow algorithm for an Ishikawa model with $N$ levels and therefore $N+1$ labels, yield:
\begin{algorithm}[h!]
$\forall L_i, i \in {0..N}, \, u_{L_i}(x) \gets 1 /  (N+1) $\;
\While{not converged}{
  $d_{L_0}(x) \gets 0$\;
  \For{$L_i$ in order $i \in [1..N]$}{
  	$d_{L_i}(x) \gets d_{L_{i-1}}(x) + \dvg q_{L_i}(x)$\;
  }
  $\forall L_i, \, d_{L_i}(x) \gets d_{L_i}(x) + D_{L_i}$\;
  
  $\forall L_i , \, u_{L_i}(x) \gets u_{L_i}(x) \exp \left( - \frac{d_{L_i}(x) }{c} \right) $\;
  $\forall L_i , \, d_{L_i}(x) \gets u_{L_i}(x) $\;
  $a(x) \gets \sum_{\forall L_i}u_{L_i}(x)$\;
  $\forall L_i, \, u_{L_i}(x) \gets u_{L_i}(x) / a(x)$\;
  
  \For{$L_i$ in order $i \in [N-1 .. 1]$}{
  	$d_{L_i}(x) \gets d_{L_i}(x) + d_{L_{i+1}}(x)'$\;
  }
  $\forall L_i, i \in {1..N}, \, q_{L_i}(x) \gets \operatorname{Proj}_{|q_{L_i}(x)| \leq S_{L_i}(x)} \left( q_{L_i}(x) - c\tau \nabla d_{L_i}(x) \right)$ \;
}
\caption{Pseudo-Flow Algorithm: Continuous Ishikawa Model}
\end{algorithm}
\newline
This algorithm involves $N+1$ buffers for the labelings, $1$ for an accumulator, $N$ for flow excess values (noting that $d_{L_0}(x)$ is largely a mathematical formality and always takes on a value of $0$ during propagation or $D_{L_0}(x)$ during label updates) and $3N$ for spatial flows. In total, this yields $5N+2$ buffers which is an improvement over the $6N+1$ buffers required to efficiently implement the algorithm Bae et al. \cite{bae_fast_2011}

\section{Discussion and Conclusions}
In this paper, we have presented a framework for the incorporation of entropic distance and Bregman proximal projections into the optimization of the Potts model. This approach can be extended to the case of Ishikawa models, hierarchical models, and general set-theoretic models with minimal additional theoretical development, demonstrating its general applicability in continuous max-flow segmentation.

The pseudo-flow approach has the theoretical advantage over the full-flow model in that it constrains intermediate segmentations to be feasible allowing for meaningful truncation. It has the practical advantage of requiring a significantly smaller memory footprint that makes it more amenable to implementation on massively parallel computing graphics hardware where memory is available in as large a quantity.

\acknowledgments 
The authors would like to acknowledge Dr.\ Elvis Chen and Jonathan McLeod for their invaluable discussion, editing, and technical support.


\bibliographystyle{spiebib}   
\bibliography{TechReportPseudoFlow} 

\end{document}